# 利用語言標記來計算中介 CTC 損失以增強語碼轉換語音辨識

# Leveraging Language ID to Calculate Intermediate CTC Loss for Enhanced Code-Switching Speech Recognition

楊子霆 Tzu-Ting Yang[1]，王馨偉 Hsin-Wei Wang[2]，陳柏琳 Berlin Chen[3]

[1, 2, 3] 國立臺灣師範大學資訊工程學系

**Computer Science and Information Engineering, National Taiwan Normal University**

E-mail: {[1]tzutingyang, [2]hsinweiwang, [3]berlin}@ntnu.edu.tw

## 摘要

近年來，端到端(End-to-End)語音辨識技術嶄露頭角，整合了傳統自動語音辨識（Automatic Speech Recognition, ASR）模型中的聲學、發音辭典和語言模型等元件。不需要事先建立發音辭典，便能達到與人類相當的辨識效果。在全球文化交流日益頻繁的背景下，多語者在對話中經常會切換不同的語言，這種被稱為語碼轉換的現象，已經成為自動語音辨識領域中不可忽視的挑戰之一。然而，由於語碼轉換的訓練資料相對稀缺，因此在遭遇到這種情況時，ASR 模型的性能往往急劇下降。過去的研究大多將語碼轉換任務拆分為多個處理單一語言的任務，然後分別學習不同語言的特定領域知識，用以簡化模型的學習複雜度。然而，這些方法極有可能忽略語言之間的交替切換機制一類的重要信息。因此，本文嘗試在 ASR 模型的編碼器中間層引入語言識別資訊，期望透過較為隱含的方式來生成蘊含語種區別的聲學特徵，從而降低模型在處理語種轉換時的混淆度。

**關鍵詞**：自動語音辨識、語碼轉換、中介損失

## Abstract

In recent years, end-to-end speech recognition has emerged as a technology that integrates the acoustic, pronunciation dictionary, and language model components of the traditional Automatic Speech Recognition (ASR) model. It is possible to achieve human-like recognition without the need to build a pronunciation dictionary in advance. However, due to the relative scarcity of training data on code-switching, the performance of ASR models tends to degrade drastically when encountering this phenomenon. Most past studies have simplified the learning complexity of the model by splitting the code-switching task into multiple tasks dealing with a single language and then learning the domain-specific knowledge of each language separately. Therefore, in this paper, we attempt to introduce language identification information into the middle layer of the ASR model's encoder. We aim to generate acoustic features that imply language distinctions in a more implicit way, reducing the model's confusion when dealing with language switching.

**Keywords**：*Automatic Speech Recognition, Code-Switching, Intermediate Loss*

表 1. 語碼轉換類型示例

|  | 示例 |
|---|---|
| 原句 | 快來不及了坐計程車去吧! |
| 句間轉換 | We're too late. 坐計程車去吧! |
| 句內轉換 | 快來不及了 get into the taxi! |

## 1. 前言

隨著文化全球化的推進，儘管世界上僅有不到四分之一的國家訂立第二母語，但現今全球已有超過六成的人口為多語者(Multilingualism)。語碼轉換(Code-Switching)現象經常不自覺的發生在多語者的對話內容當中，此一現象可進一步被細分為跨句運用不同語種的句內轉換(Intra-Sentential Code-Switching)以及不同語句間交替使用相異語言種類的句間轉換(Inter-Sentential Code-Switching)。由於句內轉換的交錯規律近乎無跡可尋、因人而異，造成這類資料在辨識時尤為艱鉅。由此可知，語碼轉換現象勢必成為自動語音辨識(Automatic Speech Recognition)在實務應用時應被考量的情境之一。

近年來端到端(End-to-End)語音辨識技術蓬勃發展，這一系列技術整合了傳統語音辨識模型中需人工建立的發音辭典(Pronunciation Lexicon)、聲學模型 (Acoustic Model) 以及語言模型 (Language Model)等部件，此舉大幅簡化了神經網路建模的難易度。透過共同最佳化的訓練方式，端到端辨識模型最大程度的降低了聲學／語言模型的不一致性，更在效果上超越傳統的語音辨識模型。

語碼轉換語料對比單語種語料顯得十分稀少，對此學者們紛紛提出許多數據增量的方法企圖減緩此一問題，可大致可分為文字／音訊兩類：在文字數據增量方面，早期學者嘗試使用語言學上的語碼轉換文法規範，如：等價性限制(Equivalence Constraint)以及功能性頭部限制(Functional Head Constraint)等產生額外的訓練文本。隨後[1]先利用機器翻譯模型取得相異語種的譯文，再針對初始的文本進行單詞/片語的替換，以此取得語碼轉換文本；在音訊方面，語碼轉換文本可經由文字轉語音(Text-to-Speech, TTS)系統合成出成對的標認資料。另外一個有效的做法是直接利用大量的單語資料參與訓練，增加單單詞涵蓋率，並使模型可以更專注於學習各自語言領域的句法資訊，[2]在模型轉錄期間將語碼轉換語料分割為相互獨立的單語種資料而後整合，期望藉由簡化任務模式來降低模型學習的難度，[3]在解碼器端根據語言種類預先遮罩後

透過自注意力機制，單獨學習各個語言種類的領域知識。

儘管各類文獻證明分化處理不同語言，可有效降低語言之間的困惑度，但分別處理語碼轉中的相異語言將喪失跨語種的上下文意資訊。因此本文嘗試在不分割語言的情況下，在編碼器的中間層引入語言識別 (language Identification, LID)資訊，期望可以在產生聲學特徵時，隱諱地迫使編碼器考量語種間的文意資訊和結構。

## 2. 相關研究

### 2.1. 連結時序分類

由於輸入的聲學表示 $\mathbf{X} = (\mathbf{x}_l \in \mathbb{R}^D | l = 1, \cdots, L)$ 與令牌層級的目標輸出序列 $\mathbf{y} = (y_s \in V | s = 1, \cdots, S)$ 之間，在大多數的情況下分別代表兩者長度的 L 與 S 長度相差極大，故而連結時序分類 (Connectionist Temporal Classification, CTC)扮演著將$\mathbf{X}$對齊至$\mathbf{y}$的關鍵角色。CTC 所輸出之似然$P_{CTC}(\mathbf{y}|\mathbf{X})$則代表了模型所產生的輸出能夠正確對齊目標令牌序列的機率，用以判斷模型的對齊能力：

$$P_{CTC}(\mathbf{y}|\mathbf{X}) = \sum_{\mathbf{a} \in A} P(\mathbf{a}|\mathbf{X}), \qquad (1)$$

其中，集合 A 包含了所有可能的對齊序列。隨後進一步以此結果作為 CTC 損失$L_{CTC}$的計算依據：

$$L_{CTC} = -log P_{CTC}(\mathbf{y}|\mathbf{X}), \qquad (2)$$

### 2.2. 骨幹網路

我們選擇 Transformer block [4] 作為編碼器模組，音訊表示在記錄位置編碼後，將被傳送到由數個編碼器模塊組成的編碼器，編碼器模塊以自注意力機制為核心元件，具體的模塊結構依序如下：

$$\mathbf{X}_{mid} = \mathbf{X}_{pre} + SelfAttention(\mathbf{X}_{pre}), \qquad (3)$$

$$\mathbf{X}_{post} = \mathbf{X}_{mid} + FeedForward(\mathbf{X}_{mid}), \qquad (4)$$

其中，$\mathbf{X}_{pre}$為前一編碼器模塊所輸出之特徵，$\mathbf{X}_{mid}$及$\mathbf{X}_{post}$則分別為中間處理的過程及編碼器模塊的輸出。如此一來，編碼器便可透過自注意力機制全

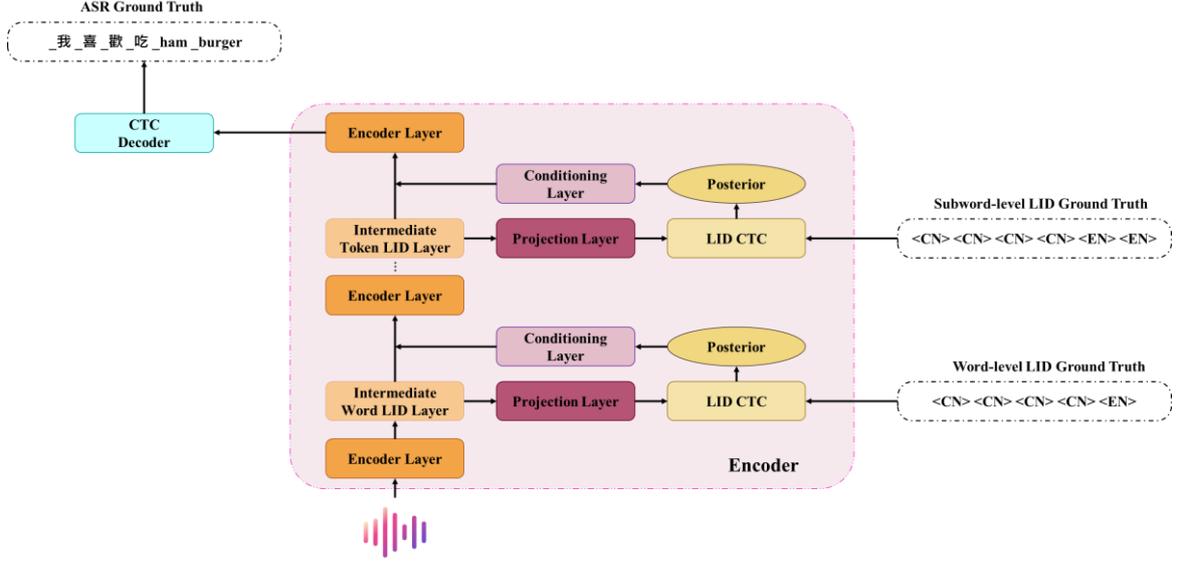

**圖 1. 語言感知編碼器模塊**

**表 2. 句內轉換語碼轉換之各層級語言識別(LID)標記**

|  | 令牌 | 語言識別標記 |
|---|---|---|
| **語句層級** | <我喜歡吃 hamburger> | <CS> |
| **單詞層級** | <我><喜><歡><吃><hamburger> | <CN> <CN> <CN> <CN> <EN> |
| **子詞層級** | <我><喜><歡><吃><ham> <burger> | <CN> <CN> <CN> <CN> <EN> <EN> |

局考量聲學特徵，建構出隱含上下文義的輸出特徵。最後經由 CTC 模組對齊後計算損失，即可對編碼器輸出進行導引。

## 3. 方法

### 3.1. 中間層連結時序分類

我們的中介語言識別編碼器(Encoder)模塊如圖一所示。若僅針對利用編碼器輸出 **H** 所計算的幀級令牌似然計算 CTC 損失，將難以規範各個重複的編碼器模塊所著重編碼的特徵。反向傳播的機制使然，造成 CTC 損失對於淺層的編碼器模組的約束力度相對較輕。同時因為淺層的編碼器模組處在聲學表示編碼的初期，在處理聲學表示時，語義資訊仍為混亂不堪的，約束力的降低可能造成模型學習效率的較差。因此[5]首度將中間層損失(Intermediate Loss)引入至編碼器內，概念類似於將多任務學習架構階層式的建構在單一編碼器內，選定 $K$ 個特定編碼器模塊的中間層(Intermediate LID Layer)輸出 $\mathbf{H}_k$

與不同類粒度的目標序列 $\mathbf{y}_g$ 利用 $CTC$ 計算中介損失(Intermediate Loss) $L_{Inter}$：

$$P_k^{Inter}(\mathbf{y}_g|\mathbf{X}) = CTC(\mathbf{y}_g, \mathbf{H}_k), \qquad (5)$$

$$L_{Inter} = -\frac{1}{|K|}\sum_{k \in K} log P_k^{Inter}(\mathbf{y}_g|\mathbf{X}), \qquad (6)$$

隨後將中介損失整合入整體損失 $L_{Total}$ 中，。

$$L_{Total} = \frac{1}{2} * (L_{CTC} + L_{Inter}). \qquad (7)$$

憑藉此一輔助限制損失，可迫使編碼器內部觸發數個子模型促使淺層編碼器模組運作。在後續的研究中，學者將訓練時將中間層 CTC 產生的似然，進一步代入下一模塊作為其編碼依據[6]，經過多次的模塊迭代後，可有效強健幀級聲學特徵中，幀與幀之間的關聯性，降低獨立假設所造成的效能減損，其作用類似於 Mask CTC[7]、Align Refine[8]：

$$\mathbf{H}_k^{post} = \mathbf{H}_k^{pre} + CL(P_k^{Inter}(\mathbf{y}_g|\mathbf{H}_k^{pre})) \qquad (8)$$

其中 $P_k^{Inter}$ 透過調節層(Conditioning Layer, $CL$)轉換

後，與原先編碼器模塊的中間層輸出 $\mathbf{H}_k^{pre}$ 相加作為下一編碼器模塊之輸入 $\mathbf{H}_k^{post}$。

先前的研究中，儘管可能在相異的編碼器模組運用不同顆粒度的標籤輔助學習，但從始至終皆以預測出目標序列 **y** 為目的[6][9]。然而我們運用語言識別資訊的目的在於輔助目標序列的預測。若單是為了語言識別資訊在編碼器內建構一個子模型，可能導致剩餘的編碼器模塊不足以針對目標序列完整的編碼。因此我們基於[10]的模型進行調整，引入語言識別映射層將編碼器中間層所輸出的聲學特徵先經過投影層(Projection Layer)投影後再進行 CTC 損失的計算。如此一來，既可以隱晦的訓練中間層的編碼器模塊對於語言的敏感度，同時大幅減少了對聲學表示編碼進程的影響。

### 3.2. 多層次語言識別標記

先前的研究將中介損失引入多語言的語音辨識中，利用 LID 協助進行語句層級的語種分類，以最小參數量取得了有效的進步[10]。在語碼轉換中，儘管語言種類完全不同，但在語言之間仍存在大量的同音異字，這些發音相似的字在傳統編碼器所產生的聲學特徵可能極為雷同。為此，我們同樣希望藉由 LID 資訊讓模型在編碼器能夠盡早的理解到語言間的差異性，以產生蘊含語言差異性的聲學特徵。這將幫助模型在後續的步驟中(如：解碼器)，可以更專注在學習特定語言的領域知識上，同時降低跨語言的文義混淆。

鑑於語碼轉換的語料在辨識難度上遠超語句層級的多語種辨識，我們希望了解 LID 在輔助辨識上可以取得多大的效能進步。我們將不同顆粒度的 LID 資訊依據難易度進行分級，依據複雜度由簡至難分為：語句層級、詞層級、子詞層級。由於語碼轉換語料中亦包含了部分單語的語句層級資料，我們將語句層級的單語和語碼轉換語料分別標記為<Monolingual>及<Code_Switch>；詞層級則依據中文字元和英語詞的單位來分割原有標記，並給予對應的中文字元標記<Chinese>以及英語單詞標記<English>；最後在子詞層級中我們承接單詞層級，進一步統計英語字根的詞頻，將英語單詞細分

### 表 3. 以 MER 表示之語言識別映射層剝離分析

| - | 不包含 語言識別映射層 | | 包含 語言識別映射層 | |
|---|---|---|---|---|
| | **Dev<sub>MAN</sub>** | **Dev<sub>SGE</sub>** | **Dev<sub>MAN</sub>** | **Dev<sub>SGE</sub>** |
| 語句層級 | 22.2 | **30.3** | **21.8** | 30.4 |
| 單詞層級 | 21.9 | 30.8 | **21.6** | **30.3** |
| 子詞層級 | 21.8 | 30.7 | **21.7** | 30.7 |

### 表 4. 相異顆粒度交錯結合之辨識效能，結果以 MER 表示

| - | 包含 語言識別映射層 | |
|---|---|---|
| | **Dev<sub>MAN</sub>** | **Dev<sub>SGE</sub>** |
| 基線模型 | 21.9 | 30.8 |
| 語句+單詞 | 23.2 | 32.2 |
| 語句+子詞 | 23.5 | 32.2 |
| 單詞+子詞 | **21.6** | **30.0** |

為多個子詞，並根據令牌的語言種類分別給予<Chinese>/<English>標記，具體示例可參見表2。

## 4. 實驗

### 4.1. 資料集

為了驗證方法的有效性，我們採用東南亞自發性語碼轉換語料庫 SEAME[11]進行訓練。SEAME 包含了約115小時的日常對話以及面試錄音，我們將其分割為98小時的訓練集、5小時的驗證集以及分別包含8小時和4小時的測試集 Dev<sub>MAN</sub> 及 Dev<sub>SGE</sub>。不同於東亞地區的語碼轉換多以借字(Borrow Word)的型式出現，且由於東南亞地區人口傾向在單一語句間更頻繁的交替使用不同語言，致使 SEAME 擁有更嚴峻的語碼轉換現象，為一兼具代表性語挑戰性的語料集。

### 4.2. 實驗設定

我們利用 ESPNET [12]來建構後續實驗中的所有模型，並且採用了 Transformer CTC 作為基線模型。具體來說，我們在基線模型上使用了12個

Transformer 模塊構成編碼器，並以 CTC 作為解碼器。我們將隱藏單元數、注意力頭、線性單位數分別設置為256、4、2048。隨後選定了 adam 作為模型的優化器以及 Warmup LR 策略，總共訓練了100個週期。所有的目標文本在進入模型前將被對應轉換為2624個中文字符、3000個英文的子詞以及3.2提及的各類 LID 標籤。在中間層模塊的選定上，為了確保模型擁有足夠多數量來應對我們的語言識別標記的學習，我們以3作為間隔，進行特定語言識別編碼器模塊的選定。例如：在同時引入語句層級和詞級的語言識別標記時，它們將分別規範第3及第6個編碼器模塊。為了突顯我們的模型在資料量稀少時的適應性，後續實驗將排除任何的資料增量方法以及語言模型的介入。訓練結束後，我們將會對最佳的前5個模型檢查點做參數平均，並以此作為最終參與測試的模型。隨後在測試期間 Beam Search 頻帶的寬度為10。最終的效能評估中，我們採用合併計算中文字符以及英文單詞錯誤數所計算出的混合錯誤率(Mixed Error Rate, MER)作為評估依據。

### 4.3. 實驗結果

首先我們可以從表3.看出，大部分包含語言識別映射層的方法在兩個測試集Dev_MAN和Dev_SGE上均有所進步。這映證了我們的猜想－模型因為調用前半段的編碼器模塊來專門處理語言識別資訊，將導致處理文字序列的可用模塊被減少。

隨後，我們透過階層式引導，希望可以誘導模型漸進式的學習語言之間的差異性。透過表4.可以看出，在單詞和子詞分別約束第3及第6個編碼器模塊時，模型可以擁有最好的效能。然而，在語句層級與其他難易度的標籤結合時，模型的效能卻急遽下降，我們推測這是因為語句層級僅包含一個 LID 標籤<Monolingual>或<Code_Switch>，造成 CTC 在對齊時的難度提升。

## 5. 結論

在本文中我們嘗試將語音識別資訊引入端到端語碼轉換語音辨識中，透過對編碼器隱晦的語言種類暗示，並改善了傳統中介損失的弊端，在後續實驗中以最少量的參數實現效能的改進。未來我們將積極嘗試更多元的識別標記組合，並將其與更強大的模型整合，以驗證方法的泛用性。

## 6. 參考文獻